\documentclass[10pt,twocolumn,letterpaper]{article}

\usepackage[pagenumbers]{cvpr} % To force page numbers, e.g. for an arXiv version

\usepackage{graphicx}
\usepackage{amsmath}
\usepackage{amssymb}
\usepackage{booktabs}

\usepackage[pagebackref,breaklinks,colorlinks]{hyperref}

\usepackage[capitalize]{cleveref}
\crefname{section}{Sec.}{Secs.}
\Crefname{section}{Section}{Sections}
\Crefname{table}{Table}{Tables}
\crefname{table}{Tab.}{Tabs.}

\def\cvprPaperID{*****} % *** Enter the CVPR Paper ID here
\def\confName{CVPR}
\def\confYear{2022}

\begin{document}

\title{Non-Hierarchical Transformers for Pedestrian Segmentation}

\author{Amani Kiruga \hspace{2em} Xi Peng \\
Deep-REAL \\
University of Delaware\\
{\tt\small \{akiruga,xipeng\}@udel.edu}
% \and
}
\maketitle

%%%%%%%%% ABSTRACT
\begin{abstract}
   We propose a methodology to address the challenge of instance segmentation in autonomous systems, specifically targeting accessibility and inclusivity. Our approach utilizes a non-hierarchical Vision Transformer variant, EVA-02, combined with a Cascade Mask R-CNN mask head. Through fine-tuning on the AVA instance segmentation challenge dataset, we achieved a promising mean Average Precision (mAP) of 52.68\% on the test set. Our results demonstrate the efficacy of ViT-based architectures in enhancing vision capabilities and accommodating the unique needs of individuals with disabilities. 

\end{abstract}

%%%%%%%%% BODY TEXT
\section{Introduction}
\label{sec:intro}

The advancement of autonomous systems, powered by robotics and artificial intelligence, has revolutionized various aspects of our lives in recent years. Autonomous systems, which range from self-driving cars and drones to smart home devices and personal assistants, provide unparalleled levels of efficiency, convenience, and adaptability. However, as these autonomous systems become increasingly prevalent, it is crucial to address the problems they present when interacting with people with disabilities. 

Pedestrian detection technologies are a fundamental component of autonomous systems, enabling them to identify and track pedestrians in real-time to avoid collisions and ensure safe navigation. While these technologies have shown promising results in general scenarios, they often fall short when it comes to accommodating the unique physical, sensory, and cognitive needs of people with disabilities. These significant hurdles must be overcome to ensure
inclusivity and equal accessibility to all.

We aim to address these issues by participating in the CVPR2023 AVA: Accessibility, Vision, and Autonomy Challenge. Through a synthetic instance segmentation and pose detection benchmark, the AVA challenge allows researchers in accessibility, computer vision, and robotics to collaboratively identify challenges and explore solutions associated with developing data-driven vision-based accessibility systems. In this work, we participate in the instance segmentation track of the challenge which contains difficult accessibility-related object categories such as 'wheelchair' and 'cane' set in a synthetic dataset called X-world \cite{Zhang2021XWorld}.

There are several challenges we encountered while training our models. Firstly, the instances belonging to the accessibility-related object categories only consist of approximately 18\% of the total object instances. Therefore the dataset has a long-tailed distribution. Secondly, given the dataset is synthetic, models pre-trained on real images may not transfer perfectly to the dataset. 

Recent work has shown that plain non-hierarchical Vision Transformer (ViT) backbones serve as reliable feature extractors for image-related downstream tasks \cite{Li2022Exploring}. In this work, we adopt a large pre-trained plain ViT variant with a Cascade Mask R-CNN head to tackle the instance segmentation AVA challenge. We use the large pre-trained EVA-02 ViT variant \cite{fang2023eva02} which is pre-trained using Masked Image Modelling (MIM) on a large merged dataset consisting of Imagenet-21K, Conceptual 12M, Conceptual 3M, COCO training set, ADE20K training set, Object365, and OpenImages for a total of 38 million images. Our method achieves a respectable mean Average Precision of 52.68\% on the CVPR2023 AVA: Accessibility, Vision, and Autonomy Segmentation Challenge test dataset. 

%-------------------------------------------------------------------------
\section{Methodology}
In this section, we introduce the model we employ and both the data augmentation and training strategies we use to fine-tune on the AVA instance segmentation challenge dataset. 

\subsection{Training strategy}
\textbf{Model}: Unlike transformer backbones like Swin \cite{liu2021swin} which use a hierarchical architecture for image processing, the original Vision Transformer \cite{dosovitskiy2021image} is a plain non-hierarchical transformer that processes images as single-scaled feature maps. 

For this challenge, we select a parameter-efficient plain ViT variant called EVA-02 \cite{fang2023eva02}. At the time of its publication, EVA-02 achieved the state-of-the-art result on both COCO and LVIS instance segmentation benchmarks with only 304M parameters. Their main contribution is a strong pre-trained model that was trained using Masked Image Modelling (MIM). The segmentation head used is the capable Cascade Mask R-CNN \cite{cai2018cascade}. The strong performance on instance segmentation benchmarks, and the parameter efficiency makes the combination of EVA-02 as a feature extractor and Cascade Mask R-CNN as a mask head, well-suited for the AVA challenge. 

\textbf{Data augmentation}: Following the strategy of EVA-02 on COCO, we use large-scale jittering for data augmentation by rescaling the images between 0.1x and 2x the original image size while keeping proportions. We then crop the image to 1920 x 1920 via padding. This was found to perform better than 1536 x 1536 used in the original COCO fine-tuning. For strong perfomance we combine both the the training and validation temporal datasets for a total of 228329 images with usable annotations.  

\textbf{Training strategy}: We use a small fine-tuning learning rate of 4e-5 with a constant learning rate schedule. We also use warmup with a relative warmup length of 0.01 of the entire training schedule and an initial learning rate of 0.001x the final learning rate. We also use a batch size of 32 and AdamW optimizer with $\beta_1=0.9, \beta_2=0.999, \epsilon=1e-8$. Finally, following the fine-tuning strategy on COCO, we use Exponential Moving Average (EMA) of the model weights with a decay rate of 0.9999. We perform end-to-end training on 8 A100 GPUs.

\section{Results}
\begin{figure}[t]
  \centering
   \includegraphics[width=0.8\linewidth]{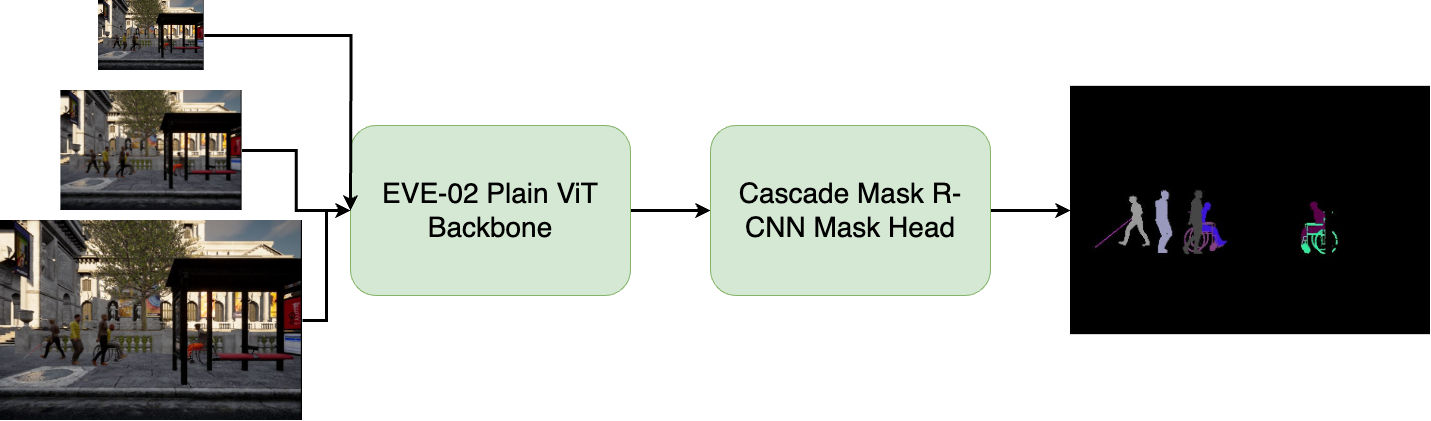}

   \caption{Our method: Large Scale Jittering, EVE-02 ViT Backbone, Cascade Mask R-CNN Mask Head.}
   \label{fig:onecol}
\end{figure}

\begin{table}
\centering
  \begin{tabular}{@{}c|c|c|c|c@{}}
    \toprule

  \multicolumn{1}{p{1.4cm}|}{\centering Training \\ Iterations}
& \multicolumn{1}{|p{1.2cm}|}{\centering Image \\ Size}
& \multicolumn{1}{|p{0.9cm}|}{\centering Batch \\ Size}
& \multicolumn{1}{|p{1.4cm}|}{\centering Learning \\ Rate}
& \multicolumn{1}{|p{1cm}}{\centering mAP \% (Val)}
\\
    \hline
    35k & $1536^2$ & 8 & 1e-5 & 42.40  \\
    35k & $1920^2$ & 8 & 1e-5 & 45.59  \\
    35k & $1920^2$ & 8 & 4e-5 & 48.49  \\
    *20.5k & $1920^2$ & 16 & 4e-5 & 49.28 \\
    \bottomrule
  \end{tabular}
  \caption{Results on the validation set for different training strategies. For all for models, the non-temporal training set is used. Longer training times, bigger learning rates, and using the original image size (with square padding) improve performance. *initialized from an 18k training iterations run with batch size 8.}
  \label{tab:results_1}
\end{table}

\begin{table}
\centering
  \begin{tabular}{@{}c|c|c|c|c@{}}
    \toprule

  \multicolumn{1}{p{1.2cm}|}{\centering Training \\ Dataset}
&  \multicolumn{1}{p{1.4cm}|}{\centering Training \\ Iterations}
& \multicolumn{1}{|p{1.2cm}|}{\centering Image \\ Size}
& \multicolumn{1}{|p{1.4cm}|}{\centering Learning \\ Rate}
& \multicolumn{1}{|p{1cm}}{\centering  mAP \% (Test)}
\\
    \hline
    Train+Val & 103k & $1920^2$ & 4e-5 & \textbf{52.68}  \\
    \bottomrule
  \end{tabular}
  \caption{Final result trained with temporal train and validation data. Model is trained with a batch size of 32. }
  \label{tab:results_2}
\end{table}

\textbf{Experiments}: We train many variants of the model to search for good hyperparameters. All experiments use the EVE-02 ViT variant with a ViT-Large backbone and Cascade Mask R-CNN mask head. See \Cref{tab:results_1}.

\textbf{Final Result}: To push our model to its full potential, we train for much longer and we switch to using the temporal data. Furthermore, we combine the train and validation sets. The model is trained with a batch size of 32 on 8 A100 GPUs. See \Cref{tab:results_2}
\section{Conclusion}
In conclusion, we present a methodology for the AVA instance segmentation challenge, focusing on accessibility and vision-based autonomy. Our approach utilizes a non-hierarchical Vision Transformer variant, EVA-02, combined with a Cascade Mask R-CNN mask head. Through extensive experimentation and fine-tuning, we achieve a promising mean Average Precision (mAP) of 52.68\% on the test set. Our results demonstrate the effectiveness of the ViT-based approach in improving accessibility and inclusivity in autonomous systems. Future work can explore further training optimizations. 

%%%%%%%%% REFERENCES
{\small
\bibliographystyle{ieee_fullname}
\bibliography{egbib}
}

\end{document}